\documentclass[10pt,twocolumn,letterpaper]{article}

\usepackage[pagenumbers]{cvpr} 

\usepackage{graphicx}
\usepackage{amsmath}
\usepackage{amssymb}
\usepackage{booktabs}
\usepackage{amsfonts}
\usepackage{bbding} 
\usepackage{color}
\usepackage{multirow}
\usepackage[accsupp]{axessibility}
\usepackage{makecell}

\usepackage[pagebackref,breaklinks,colorlinks]{hyperref}

\usepackage[capitalize]{cleveref}
\crefname{section}{Sec.}{Secs.}
\Crefname{section}{Section}{Sections}
\Crefname{table}{Table}{Tables}
\crefname{table}{Tab.}{Tabs.}

\def\cvprPaperID{13976} 
\def\confName{CVPR}
\def\confYear{2024}

\begin{document}

\title{DriveDreamer: Towards Real-world-driven World Models \\ for Autonomous Driving}

\author{Xiaofeng Wang\footnotemark[1]~\textsuperscript{\rm 1}~~Zheng Zhu\footnotemark[1]~\textsuperscript{\rm 1}\textsuperscript{\Envelope}~~Guan Huang\textsuperscript{\rm 1,2}~~Xinze Chen\textsuperscript{\rm 1}~~Jiagang Zhu\textsuperscript{\rm 1}~~Jiwen Lu\textsuperscript{\rm 2}\\
\textsuperscript{\rm 1}GigaAI
~ ~ \textsuperscript{\rm 2}Tsinghua University
\\
\small{Project Page: \url{https://drivedreamer.github.io}}
}

\twocolumn[{%
\maketitle
\begin{center}
\centering
\resizebox{0.95\linewidth}{!}{
\includegraphics{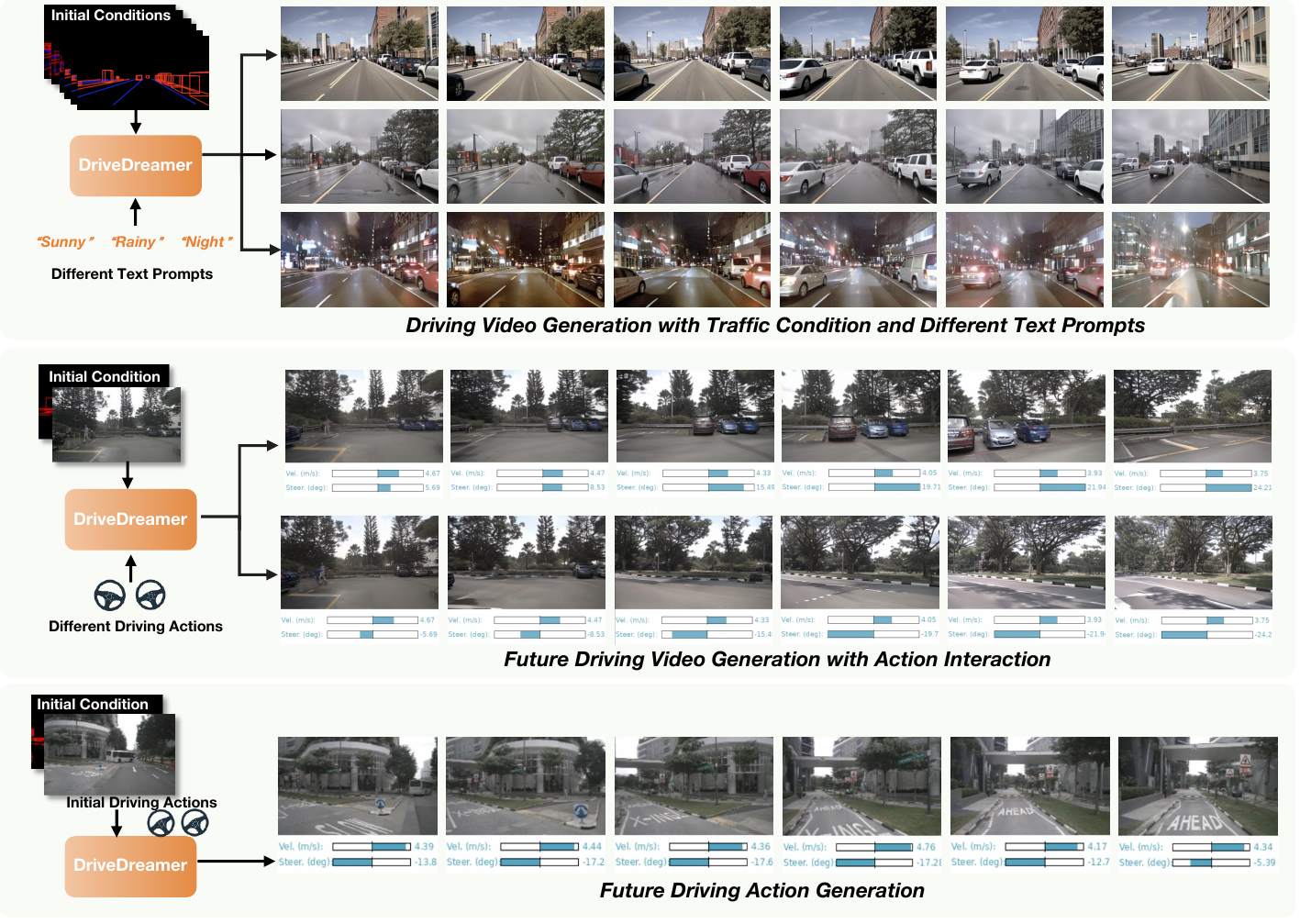}}
\captionof{figure}{\textit{DriveDreamer} demonstrates a comprehensive understanding of driving scenarios. It excels in controllable driving video generation, aligning seamlessly with text prompts and structured traffic constraints. \textit{DriveDreamer} can also interact with the driving scene and predict different future driving videos, based on input actions. Furthermore, \textit{DriveDreamer} extends its utility to anticipate future actions.}
\label{fig:fig0}
\end{center}}]


\begin{abstract}
\vspace{-1em}
World models, especially in autonomous driving, are trending and drawing extensive attention due to their capacity for comprehending driving environments. The established world model holds immense potential for the generation of high-quality driving videos, and driving policies for safe maneuvering. However, a critical limitation in relevant research lies in its predominant focus on gaming environments or simulated settings, thereby lacking the representation of real-world driving scenarios. Therefore, we introduce \textit{DriveDreamer}, a pioneering world model entirely derived from real-world driving scenarios. Regarding that modeling the world in intricate driving scenes entails an overwhelming search space, we propose harnessing the powerful diffusion model to construct a comprehensive representation of the complex environment. Furthermore, we introduce a two-stage training pipeline. In the initial phase, \textit{DriveDreamer} acquires a deep understanding of structured traffic constraints, while the subsequent stage equips it with the ability to anticipate future states. The proposed \textit{DriveDreamer} is the first world model established from real-world driving scenarios. We instantiate \textit{DriveDreamer} on the challenging nuScenes benchmark, and extensive experiments verify that \textit{DriveDreamer} empowers precise, controllable video generation that faithfully captures the structural constraints of real-world traffic scenarios.  Additionally, \textit{DriveDreamer} enables the generation of realistic and reasonable driving policies, opening avenues for interaction and practical applications.

\end{abstract}

\section{Introduction}

Spurred by insights from AGI (Artificial General Intelligence) and the principles of embodied AI, a profound transformation in autonomous driving is underway.  Autonomous vehicles rely on sophisticated systems that engage with and comprehend the real driving world. At the heart of this evolution is the integration of world models \cite{dreamv1,dreamv2,dreamv3,worldmodel}. World models hold great promise for generating diverse and realistic driving videos, encompassing even long-tail scenarios, which can be utilized to train various driving perception approaches. Furthermore, the predictive capabilities in world models facilitate end-to-end driving, ushering in a new era of autonomous driving experiences.

Deriving latent dynamics of world models from visual signals was initially introduced in video prediction \cite{dreamv3,videopred2,videopred3}. By extrapolating from observed visual sequences, video prediction methods can infer future states of the environment, effectively modeling how objects and entities within a scene will evolve over time. However, modeling the intricate driving scenarios in pixel space is challenging due to the large sampling space \cite{limit,e2esurvey}. To alleviate this problem, recent research endeavors have sought innovative strategies to enhance sampling efficiency. ISO-Dream \cite{isodream} explicitly disentangles visual dynamics into controllable and uncontrollable states. MILE \cite{mile} strategically incorporates world modeling within the Bird's Eye View (BEV) semantic segmentation space, complementing world modeling with imitation learning. SEM2 \cite{sem2} further extends the Dreamer framework into BEV segmentation maps, utilizing Reinforce Learning (RL) for training. Despite the progress witnessed in world models, a critical limitation in relevant research lies in its predominant focus on simulation environments. \looseness=-1

In this paper, we propose \textit{DriveDreamer}, which pioneers the construction of comprehensive world models from real driving videos and human driver behaviors. 
Considering the intricate nature of modeling real-world driving scenes, we introduce the Autonomous-driving Diffusion Model (\textit{Auto-DM}), which empowers the ability to create a comprehensive representation of the complex driving environment.
We propose a two-stage training pipeline. In the first stage, we train \textit{Auto-DM} by incorporating traffic structural information as intermediate conditions, which significantly enhances sampling efficiency. Consequently, \textit{Auto-DM} exhibits remarkable capabilities in comprehending real-world driving scenes, particularly concerning the dynamic foreground objects and the static background.  In the second-stage training, we establish the world model through video prediction. Specifically, driving actions are employed to iteratively update future traffic structural conditions, which enables \textit{DriveDreamer} to anticipate variations in the driving environment based on different driving strategies.
Moreover, \textit{DriveDreamer} extends its predictive prowess to foresee forthcoming driving policies, drawing from historical observations and \textit{Auto-DM} features. Thus creating a executable, and predictable driving world model.

The main contributions of this paper can be summarized as follows: (1) We introduce \textit{DriveDreamer}, which is the first world model derived from real-world driving scenarios. \textit{DriveDreamer} can jointly enable the generation of high-quality driving videos and reasonable driving policies.
(2) To enhance the comprehension of real-world driving scenes and expedite the world model convergence, we introduce the Autonomous-driving Diffusion Model and a two-stage training pipeline. The first-stage training enables the comprehension of traffic structural information, and the second-stage video prediction training empowers the predictive capacity.
(3) \textit{DriveDreamer} can controllably generate driving scene videos that are highly aligned with traffic constraints (see Fig.~\ref{fig:fig0}), enhancing the training of driving perception methods (e.g., 3D detection). Besides, \textit{DriveDreamer} can generate future driving policies based on historical observations and \textit{Auto-DM} features. Notably, \textit{DriveDreamer} achieves promising planning results in open-loop assessments on the nuScenes dataset.

\section{Related Work}

\noindent

\subsection{Diffusion Model}
Diffusion models represent a family of probabilistic generative models that progressively introduce noise to data and subsequently learn to reverse this process for the purpose of generating samples \cite{diffsurvey}. These models have recently garnered significant attention due to their exceptional performance in various applications, setting new benchmarks in image synthesis \cite{df6,vqdiff,dalle2,unidiff,glide}, video generation \cite{vdm1,vdm2,vdm3,vdm4,vdm5,vdm6}, and 3D content generation \cite{df,fantasia3d,prolificdreamer,lin2023magic3d}. To enhance the controllable generation capability, ControlNet \cite{controlnet}, GLIGEN \cite{gligen}, T2I-Adapter \cite{t2ia} and Composer \cite{composer} have been introduced to utilize various control inputs, including depth maps, segmentation maps, canny edges, and sketches. Concurrently, BEVControl \cite{bevcontrol}, MagicDrive \cite{magicdrive} and DrivingDiffuson \cite{drivediff} incorporate layout conditions to enhance image generation. The fundamental essence of diffusion-based generative models lies in their capacity to comprehend and understand the intricacies of the world. Harnessing the power of these diffusion models, \textit{DriveDreamer} seeks to comprehend the complex realm of autonomous-driving scenarios.

\subsection{Video Generation}
Video generation and video prediction are effective approaches to understanding the visual world. In the realm of video generation, several standard architectures have been employed, including Variational Autoencoders (VAEs) \cite{videopred2, vae2}, auto-regressive models \cite{ar1,ar2,ar3,ar4}, flow-based models \cite{flow1}, and Generative Adversarial Networks (GANs) \cite{gan1,gan2,gan3,gan4}. Recently, the burgeoning diffusion models \cite{df1,df2,df3,glide,df5,df6} have also been extended to the domain of video generation. Video diffusion models \cite{vdm1,vdm2,vdm3,vdm4,vdm5,vdm6} exhibit higher-quality video generation capabilities, producing realistic frames and transitions between frames while offering enhanced controllability. They accommodate various input control conditions such as text, canny, sketch, semantic maps, and depth maps.

Video prediction models represent a specialized form of video generation models, sharing numerous similarities. In particular, video prediction involves anticipating future video changes based on historical video observations \cite{mcvd,vp1,dreamv3,videopred2,videopred3}. DriveGAN \cite{drivegan} establishes associations between driving actions and pixels, predicting future driving videos by specifying future driving policies. In contrast, \textit{DriveDreamer} incorporates structured traffic conditions, text prompts, and driving actions as inputs, empowering precise, realistic video and action generation that are faithfully aligned with real-world driving scenarios.

\noindent
\subsection{World Models}
World models have been extensively explored in model-based imitation learning, demonstrating remarkable success in various applications \cite{dreamv1,dreamv2,dreamv3,gamegan,worldmodel,wm1,wm2,wm3,wm4,wm5}. These approaches typically leverage Variational Autoencoders (VAE) \cite{vae} and Long Short-Term Memory (LSTM) \cite{lstm} to model transition dynamics and rendering functionality. World methods target at establishing dynamic models of environments, enabling agents to be predictive of the future. This aspect is of paramount importance in autonomous driving, where precise predictions about the future are essential for safe maneuvering. However, constructing world models in autonomous driving presents unique challenges, primarily due to the high sample complexity inherent in real-world driving tasks \cite{e2esurvey}.
To address these problems, ISO-Dream \cite{isodream} introduces an explicit disentanglement of visual dynamics into controllable and uncontrollable states. MILE \cite{mile} strategically incorporates world modeling within the BEV semantic segmentation space, enhancing world modeling through imitation learning. SEM2 \cite{sem2} extends the Dreamer framework into BEV segmentation maps, employing reinforcement learning for training. Despite the progress witnessed in world models, a critical limitation in relevant research lies in its predominant focus on simulation environments. The transition to real-world driving scenarios remains an under-explored frontier.

\begin{figure}[t]
\centering
\resizebox{1\linewidth}{!}{
\includegraphics{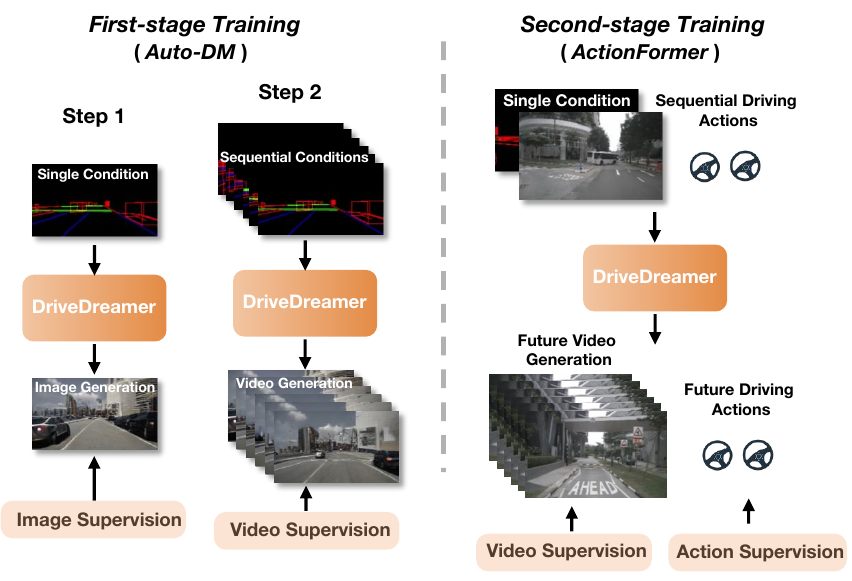}}
\caption{Two-stage training pipeline of \textit{DriveDreamer}.}
\label{fig:trainval}
\end{figure}

\begin{figure*}[ht]
\centering
\resizebox{1\linewidth}{!}{
\includegraphics{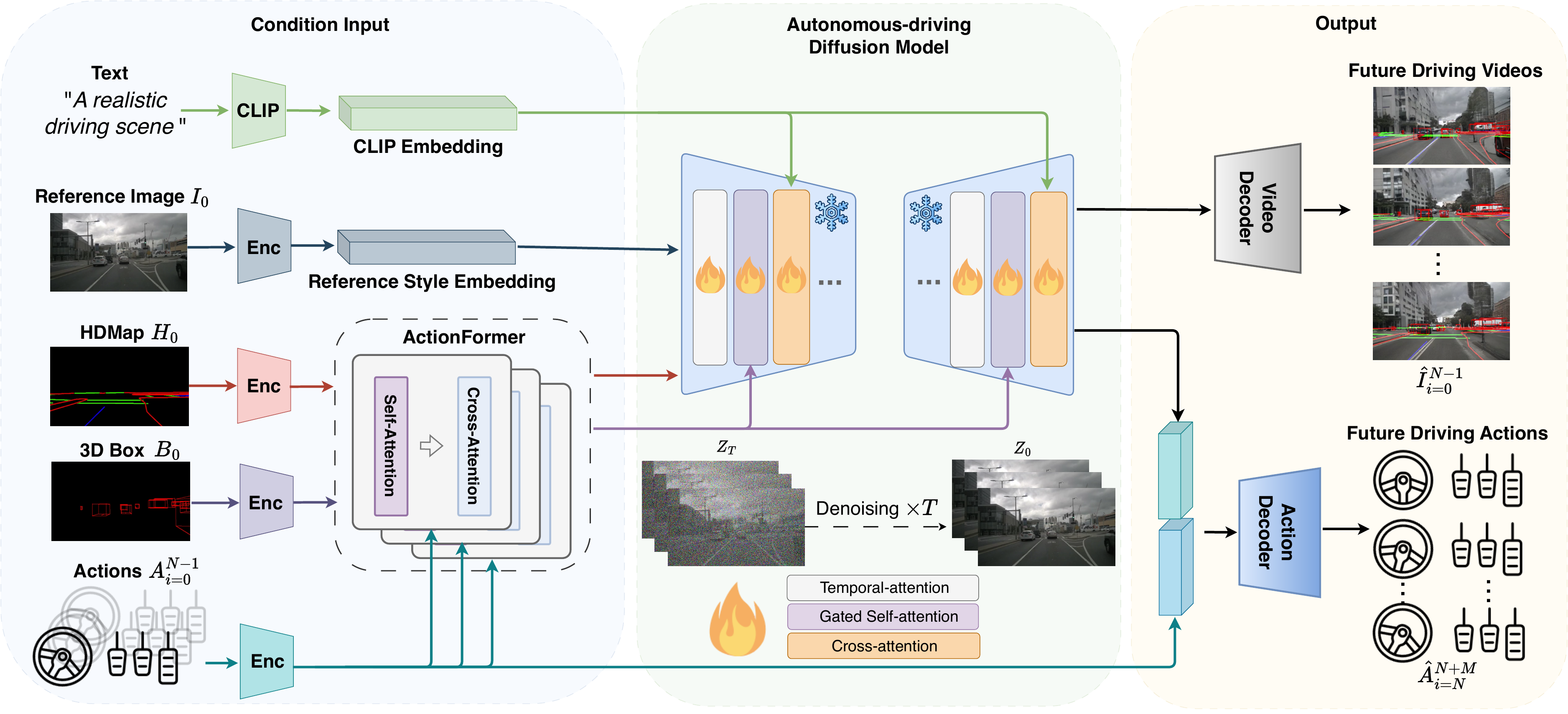}}
\caption{Overall framework of \textit{DriveDreamer}. The framework initiates with reference frame $I_0$ and road structural information (i.e., HDMap $H_0$ and 3D box $B_0$). \textit{DriveDreamer} employs the \textit{ActionFormer} to predict future road structural features, which serve as conditions provided to \textit{Auto-DM}, generating future driving videos $\hat{I}_{i=0}^{N-1}$. Additionally, text prompts enable dynamic scenario style adjustments. The model integrates past driving actions and multi-scale features from \textit{Auto-DM} to generate plausible future driving actions $\hat{A}_{i=N}^{N+M}$.}
\label{fig:drivedreamer}
\end{figure*}

\section{DriveDreamer}

The overall framework of \textit{DriveDreamer} is depicted in Fig \ref{fig:drivedreamer}. 
The framework begins with an initial reference frame $I_{0}$ and its corresponding road structural information (\ie, HDMap $H_{0}$ and 3D box $B_{0}$). Within this context, \textit{DriveDreamer} leverages the proposed \textit{ActionFormer} to predict forthcoming road structural features in the latent space. These predicted features serve as conditions and are provided to \textit{Auto-DM}, which generates future driving videos. Simultaneously, the utilization of text prompts allows for dynamic adjustments to the driving scenario style (\eg, weather and time of the day). Moreover, \textit{DriveDreamer} incorporates historical action information and the multi-scale latent features extracted from \textit{Auto-DM}, which are combined to generate reasonable future driving actions.  In essence, \textit{DriveDreamer} offers a comprehensive framework that seamlessly integrates multi-modal inputs to generate future driving videos and driving policies, thereby advancing the capabilities of autonomous-driving systems.

Regarding the extensive search space of establishing world models in real-world driving scenarios, we introduce a two-stage training strategy for \textit{DriveDreamer}. This strategy is designed to significantly enhance sampling efficiency and expedite model convergence. The two-stage training is illustrated in Fig.~\ref{fig:trainval}. There are two steps in the first-stage training. Step 1 involves utilizing the single-frame structured condition, which guides \textit{DriveDreamer} to generate driving scene image, facilitating its comprehension of structural traffic constraints. Step 2 extends its understanding into video generation.
The second-stage training enables \textit{DriveDreamer} to interact with the environment and predict future states effectively. This phase takes an initial frame image along with its corresponding structured information as input. Simultaneously, sequential driving actions are provided, with the model expected to generate future driving videos and future driving actions.
In the following sections, we delve into the specifics of the model architecture and training pipelines.

\subsection{First-stage Training}

\noindent
\textbf{\textit{Auto-DM}.}
In \textit{DriveDreamer}, we introduce \textit{Auto-DM}, to model and comprehend driving scenarios from real-world driving videos. It is noted that comprehending driving scenes solely from pixel space presents challenges due to extensive search space in real-world driving scenarios. To mitigate this, we explicitly incorporate structured traffic information as conditional inputs.

\begin{figure*}[ht]
\centering
\resizebox{1\linewidth}{!}{
\includegraphics{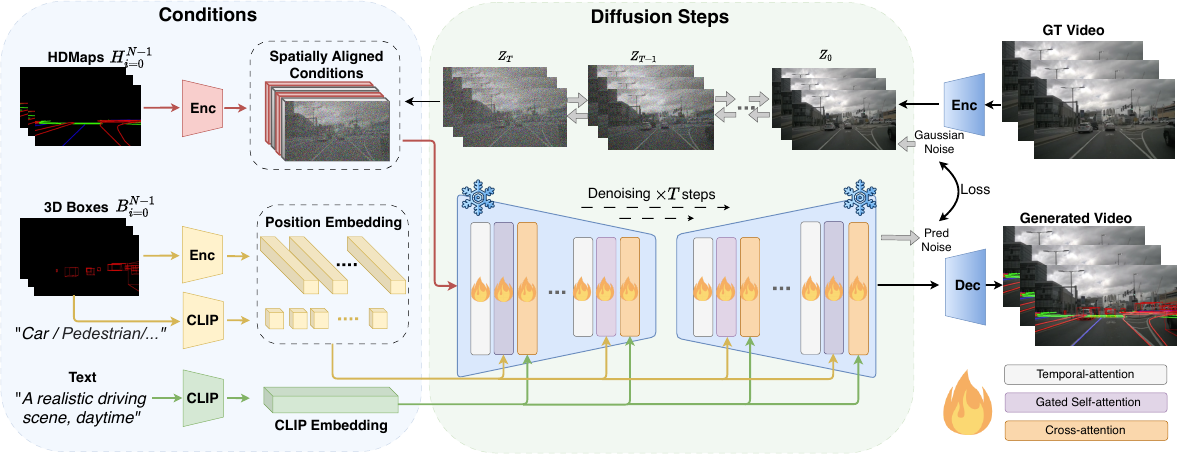}}
\caption{Overall structure of the \textit{Auto-DM}. \textit{Auto-DM} takes three types of control conditions as inputs. Spatially aligned conditions (\ie, HDMap $H_{i=0}^{N-1}$), are concatenated with noise images and fed into the diffusion steps. Position conditions, represented by 3D boxes $B_{i=0}^{N-1}$ and their labels, are flattened and utilized in the gated self-attention. Text prompts are incorporated into diffusion steps using cross-attention, influencing the style of the generated driving video. Temporal attention layers are employed to ensure the consistency of the generated video frames. The diffusion steps estimate noise and generate loss with the input noise to optimize \textit{Auto-DM}.}
\label{fig:autodm}
\end{figure*}

The overall structure of \textit{Auto-DM} is illustrated in Fig.~ \ref{fig:autodm}, where traffic conditions are projected onto the image plane, generating HDMap conditions $\{H_i\}_{i=0}^{N-1} \in \mathcal{R}^{N\times H\times W \times 3}$, and 3D boxes conditions $\{B_i\}_{i=0}^{N-1}\in\mathcal{R}^{N\times N_B\times 16}$, along with the box categories $\{C_i\}_{i=0}^{N-1}\in\mathcal{R}^{N\times N_B}$ ($N$ is the number of video frames, and $N_B$ is the predefined maximum box numbers with zero padded). In this following, unless specified, the subscript $i$ is omitted for readability.
To enable controllability, the spatially aligned conditions $H$ are encoded by convolution layers and then concatenated with $\mathcal{Z}_t$,
where $\{\mathcal{Z}_t\}_{i=0}^{N-1}$ are noisy latent features generated by the forward diffusion process \cite{df6}. For position conditions (\ie, 3D boxes) that are not spatially aligned with $\mathcal{Z}_t$, we first aggregate position embeddings $H^p$:
\begin{equation}
    H^p = \mathcal{F}_\alpha(
    [
    C_e,\text{Fourier}(B)
    ]
    ),
\end{equation}
where $\mathcal{F}_a$ is MLP layers, $C_e$ is CLIP \cite{clip} embed box categories features, $\text{Fourier}(\cdot)$ is Fourier embedding \cite{nerf}, and $[\cdot]$ is the concatenation operation. Then 
\textit{gated self-attention} \cite{gligen} is leveraged to integrate position embeddings $H^p$ with visual signals $v$ from the original UNet features \cite{df6}:
\begin{equation}
    v = v + \text{tanh}(\eta)\cdot \text{TS}(\mathcal{F}_s([v, H^p])),
\end{equation}
where $\eta$ is a learnable parameter, $\mathcal{F}_s$ is self-attention, and $\text{TS}(\cdot)$ is the token selection operation that considers visual tokens only \cite{gligen}.

To further empower \textit{Auto-DM} with comprehension of driving dynamics, we introduce temporal attention layers $\mathcal{F}_t$ to enhance frame coherence in the generated videos:
\begin{equation}
    \mathcal{F}_t(v) = \text{Reshape}(\mathcal{F}_s(\text{Reshape}(v+\mathcal{T}_{\text{pos}}))),
\end{equation}
where we first reshape the visual signal $v$ from  $\mathcal{R}^{N\times C\times H \times W}$ to $\mathcal{R}^{C\times NHW}$. The shape transformation facilitates the frame-wise self-attention layers $\mathcal{F}_s$ to learn inter-frame dynamics. $\mathcal{T}_{\text{pose}}$ denotes temporal position embeddings that are encoded by sinusoidal function \cite{videoldm}.
Finally, we restore the visual signal to its original dimensions, thus ensuring the feature integrity. Notably, the same architecture can be extended to generate multi-view images (see Fig.~\ref{fig:fig1}), where the $\mathcal{F}_s$ solely attends to neighbor views. Additionally, a stack of frame-wise attention and view-wise attention contributes to multi-view video generation (see supplement for more details). 

Furthermore, cross-attention layers \cite{df6} are utilized to facilitate feature interactions between text inputs and visual signals, empowering text descriptions to influence driving scene attributes such as weather and time of day. In the next, we will elaborate on the first-stage training pipeline, which involves two steps.

\noindent
\textbf{Step 1 training.}
The \textit{Auto-DM} incorporates input solely from a single frame of structured traffic conditions, coupled with supervision from a single-frame image.  For structured traffic conditions, HDMaps and 3D boxes are obtained either from human annotations or pertained perception methods (\eg, LAV \cite{lav}, BEVerse \cite{beverse}, UniAD \cite{uniad}). Then three-channel HDMaps (lane boundary, lane divider, and pedestrian crossing) and eight-corner 3D boxes are projected onto the image plane to generate corresponding conditions. Notably, during step 1 training, temporal attention layers are omitted, which enables the network to focus exclusively on learning the traffic structural constraints, expediting the convergence of the training process.

 \noindent
\textbf{Step 2 training.} 
The \textit{Auto-DM} incorporates input from multiple frames of structured traffic conditions and is supervised using driving videos. In contrast to step 1, learning from videos allows \textit{Auto-DM} to gain a deeper understanding of the intricate motion transitions in driving scenarios.
Building upon the pretrained models established in step 1, step 2 incorporates temporal attention layers into the model architecture. These additional parameters enable the \textit{Auto-DM} to focus on the temporal dynamics present in the input data, further enhancing its ability to capture and interpret the nuanced temporal aspects of driving scenes.

\begin{figure*}[ht]
\centering
\resizebox{1\linewidth}{!}{
\includegraphics{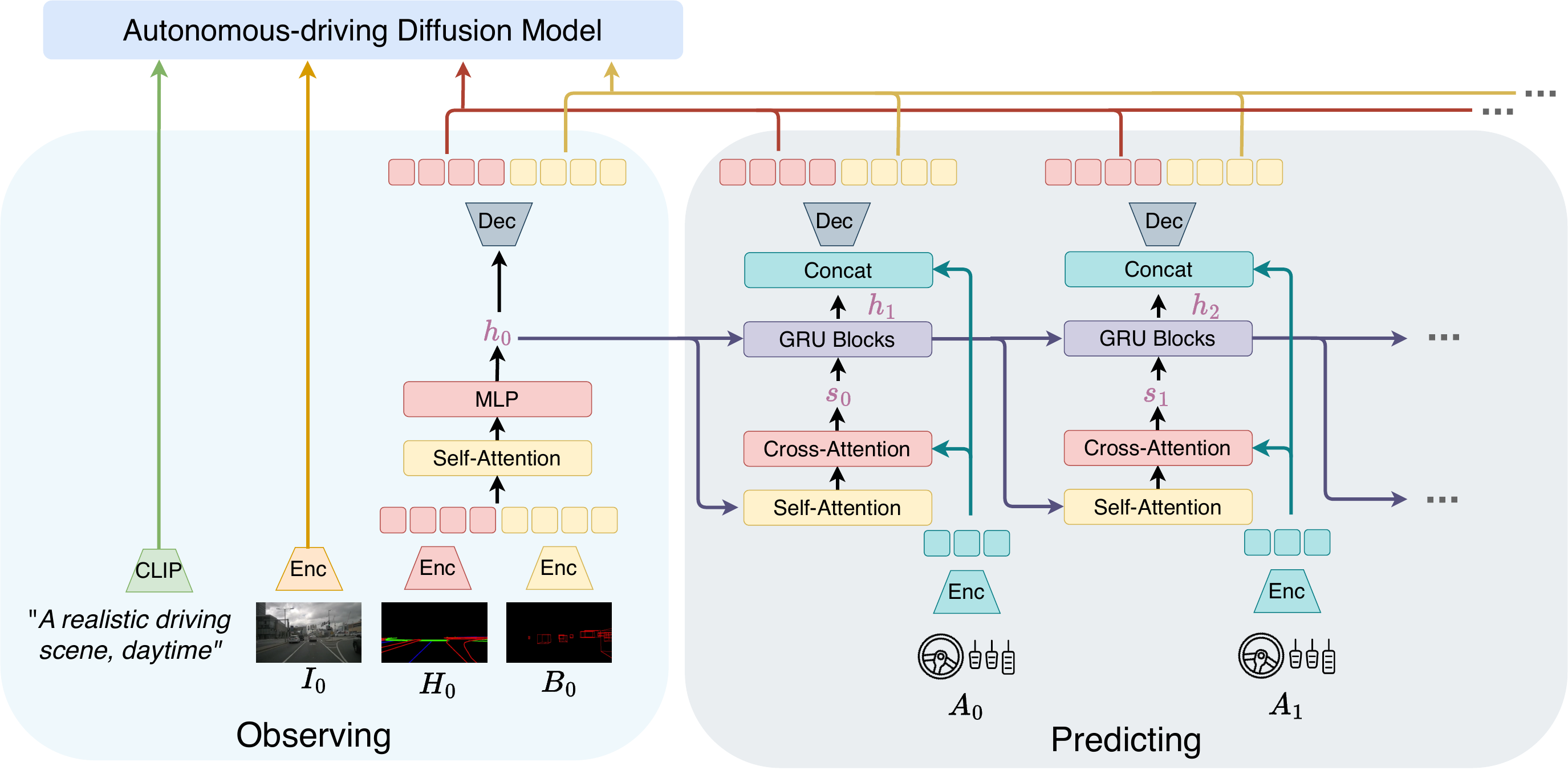}}
\caption{Overall structure of \textit{ActionFormer}. The initial structural conditions $H_0$ and $B_0$ are first encoded and flattened into a 1D latent space. These latent features are then concatenated and processed through self-attention and MLP layers, generating the hidden state. Cross-attention layers establish associations between hidden states and driving actions. Gated Recurrent Units (GRUs) are employed to iteratively predict future hidden states. These predicted hidden states are further concatenated with action features and decoded into future traffic structural conditions to be fed into \textit{Auto-DM}.}
\label{fig:actionformer}
\end{figure*}

In step 1 and step 2 training, the proposed \textit{Auto-DM} is trained using the same noise schedule as the underlying image model \cite{df6}. Specifically, the forward process gradually adds noise $\epsilon$ to the latent feature $\mathcal{Z}_0$, resulting in the noisy latent feature $\mathcal{Z}_T$. Then we
train $\epsilon_{\phi}$ to predict the noise we added, and the trainable parameters $\phi$ are optimized via:
\begin{equation}
\min _{\phi} \mathcal{L}=\mathbb{E}_{\mathcal{Z}_{0}, \epsilon \sim \mathcal{N}(\mathbf{0}, \mathbf{I}), t, c}\left[\left\|\epsilon-\epsilon_{\phi}\left(\mathcal{Z}_{t}, t, c\right)\right\|_{2}^{2}\right],
\end{equation}
where $\phi$ denotes the trainable parameters involved in the gated self-attention, temporal attention, and cross-attention layers, and time step $t$ is uniformly sampled from $[1,T]$.

\subsection{Second-stage Training}

Based on the first-stage training, \textit{DriveDreamer} has obtained comprehension of the structured traffic information. However, the desired world model should also be predictive of the future and can interact with the environment.
Therefore, we embark on the second phase of our approach. In this phase, we leverage the video prediction task to establish the driving world model. Specifically, the video prediction task entails providing an initial observation $I_0, H_0, B_0$, as well as driving actions $\{A_i\}_{i=0}^{T-1}$, with the desired outcome being the future driving videos $\{I_i\}_{i=1}^{T}$, and future driving actions $\{A_i\}_{i=T}^{T+N}$.

\noindent
\textbf{\textit{ActionFormer}.}
Recall that the trained \textit{Auto-DM} can generate driving videos $\{I_i\}_{i=0}^{T}$ based on sequential structured information $\{H_i\}_{i=0}^{T}, \{B_i\}_{i=0}^{T}$. However, in the video prediction task, future traffic structural conditions beyond the present timestamp is unavailable. To address this challenge, we introduce the \textit{ActionFormer}, which leverages driving actions $\{A_i\}_{i=0}^{T-1}$ to iteratively predict future structural conditions. The overall architecture of \textit{ActionFormer} is in Fig.~\ref{fig:actionformer}. Firstly the initial structural conditions $H_0,B_0$ are encoded and flattened into 1D latent space. The latent features are concatenated and aggregated by self-attention and MLP layers to generate the hidden state $\mathbf{h}_0$. Subsequently, cross-attention layers $\mathcal{F}_{ca}$ are utilized to construct associations between hidden states and driving actions. Then latent variable $\mathbf{s}_t$ is parameterized as:
\begin{equation}
    \mathbf{s}_{t} \sim \mathcal{N}\left(\mu_{\theta}\left(\mathcal{F}_{ca}(\mathbf{h}_t, A_t)\right), \sigma_{\theta}\left(\mathcal{F}_{ca}(\mathbf{h}_t, A_t)\right) \boldsymbol{I}\right),
\end{equation}
where $\mu_{\theta}, \sigma_{\theta}$ are layers to learn Gaussian parameters. To predict future hidden states, we employ Gated Recurrent Units (GRUs) to iteratively make updates:
\begin{equation}
    \mathbf{h}_{t+1} = \mathcal{F}_\text{GRU}(\mathbf{h}_t, \mathbf{s}_t).
    \label{eq:hidden}
\end{equation}
These hidden states are concatenated with action features and are decoded into future traffic structural conditions.
It's noted that the \textit{Actionformer} forecasts future traffic conditions at the feature level, which mitigates noise interference at the pixel level, resulting in more robust predictions. Besides the traffic structural conditions generated by \textit{Actionformer} and the text prompt condition, we process the reference image condition $I_0$ similar to \cite{videoldm}. Based on the above conditions, we extend \textit{Auto-DM} to jointly generate future driving videos  $\{I_i\}_{i=1}^{T}$  and driving actions  $\{A_i\}_{i=T}^{T+N}$.
We formalize this process as a generative probabilistic model, where the joint probability can be factorized as:
\begin{small}
\begin{equation}
\begin{aligned}
 &p\left(I_{0: T}, A_{0: T+N}, \mathbf{h}_{0: T}, \mathbf{s}_{0:T-1} \right)\\
    &\quad=p\left(I_{1:T}, A_{T:T+N} \mid \mathbf{h}_{0:T}, \mathbf{s}_{0:T-1}, A_{0:T-1}, I_0\right)\\
    &\quad \quad \prod_{t=0}^{T} p\left(\mathbf{h}_{t}, \mathbf{s}_{t} \mid \mathbf{h}_{t-1}, \mathbf{s}_{t-1}, A_{t}\right),
\end{aligned}
\end{equation}
\end{small}
where 
\begin{small}
\begin{equation}
\begin{aligned}
&p\left(\mathbf{h}_{t}, \mathbf{s}_{t} \mid \mathbf{h}_{t-1}, \mathbf{s}_{t-1}, A_{t}\right) \\
&\quad\quad\quad=p\left(\mathbf{h}_{t} \mid \mathbf{h}_{t-1}, \mathbf{s}_{t-1}\right) p\left(\mathbf{s}_{t} \mid \mathbf{h}_{t}, A_{t}\right) \\
&p\left(I_{1:T}, A_{T:T+N} \mid \mathbf{h}_{0:T}, \mathbf{s}_{0:T-1}, A_{0:T-1}, I_0\right) \\
&\quad\quad\quad=p\left(I_{1:T}\mid \mathbf{h}_{0:T}, \mathbf{s}_{0:T-1}, A_{0:T-1}, I_0\right) \\
&\quad\quad\quad\quad\quad p\left(A_{T:T+N}\mid \mathbf{h}_{0:T}, \mathbf{s}_{0:T-1}, A_{0:T-1}, I_0\right).
\end{aligned}
\end{equation}
\end{small}

\begin{figure*}[ht]
\centering
\resizebox{1\linewidth}{!}{
\includegraphics{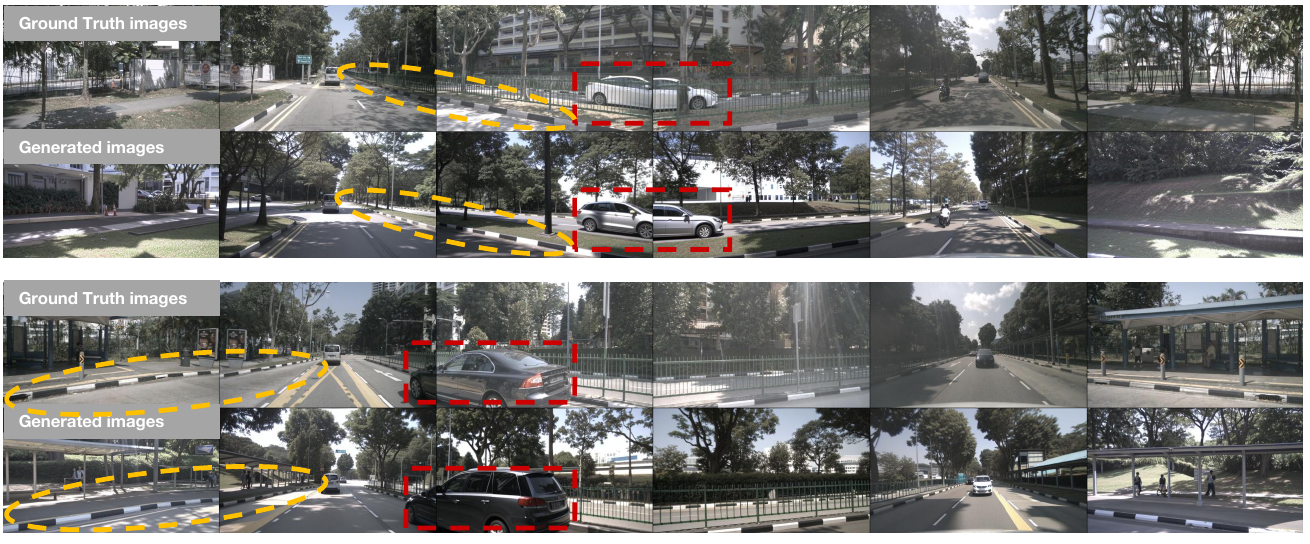}}
\caption{Visualizations of generated multi-view images, where the generation conditions (HDMaps, 3D boxes) are from nuScenes validation set. Regions highlighted by \textcolor{red}{red} rectangles and \textcolor{yellow}{yellow} circles indicate that the generated images share multi-view consistency and are aligned with ground truth conditions.}
\label{fig:fig1}
\end{figure*}

Considering updating hidden states $p(\mathbf{h}_{t} \mid \mathbf{h}_{t-1}, \mathbf{s}_{t-1})$ is a deterministic process (Eq.~\ref{eq:hidden}), only latent variables $\mathbf{s}_t$ are needed to be inferred to maximize the marginal likelihood of observation $p(I_{1:T},A_{T:T+N})$. Therefore, variational distribution $q_{\text{vd}}$ is introduced to conduct variational inference:
\begin{small}
\begin{equation}
\begin{aligned}
q_\text{vd} &\triangleq q\left(\mathbf{h}_{1: T}, \mathbf{s}_{1: T} \mid I_{0: T}, A_{0: T+N}\right)\\
&=\prod_{t=1}^{T} q\left(\mathbf{h}_{t} \mid \mathbf{h}_{t-1}, \mathbf{s}_{t-1}\right) q\left(\mathbf{s}_{t} \mid I_{\leq t}, A_{<t}\right),
\end{aligned}
\end{equation}
\end{small}

\noindent
where $q\left(\mathbf{h}_{t} \mid \mathbf{h}_{t-1}, \mathbf{s}_{t-1}\right)=p\left(\mathbf{h}_{t} \mid \mathbf{h}_{t-1}, \mathbf{s}_{t-1}\right)$. Similar to \cite{mile}, the variational lower bound can be derived as:
\begin{small}
\begin{equation}
\begin{aligned}
&\log p  \left(I_{1: T},\right. \left. A_{T: T+N}\right)\ge   \\
 &\mathbb{E}_{q\left(\mathbf{h}_{1: T}, \mathbf{s}_{1: T} \mid I_{0:T}, A_{0:T+N}\right)}[\underbrace{\log p\left(I_{1:T} \mid \mathbf{h}_{0:T}, \mathbf{s}_{0:T-1}, A_{0:T-1}, I_0\right)}_{\text {video prediction }}\\
 &\quad +\underbrace{\log p\left(A_{T:T+N} \mid \mathbf{h}_{0:T}, \mathbf{s}_{0:T-1}, A_{0:T-1}, I_0\right)}_{\text {action predcition}}].
 \label{eq:vi}
\end{aligned}
\end{equation}
\end{small}

\noindent
Note that the posterior and prior matching \cite{mile} is not included, as we empirically find the simplified variational lower bound produces similar plausible results. In Eq.~\ref{eq:vi}, the video prediction and action prediction parts can be modeled by Gaussian distributions $\mathcal{N}(\mathcal{G}(\mathbf{h}_{0:T}, A_{0:T-1},I_0), \mathbf{I})$ and Laplace distribution $\text{Laplace}(\pi(\mathbf{h}_{0:T}, A_{0:T-1},I_0), 1)$.  Therefore, we employ mean-squared error and $L_1$ loss to optimize the video prediction training. $\mathcal{G}$, $\pi$ are learnable layers involved in \textit{ActionFormer}, \textit{Auto-DM}, video decoder (\ie, VAE decoder) and action decoder. For action prediction details, we first pool multi-scale UNet features from \textit{Auto-DM}. The pooled features are concatenated with historical action features, which are then decoded by MLP layers to generate future driving actions.

Based on the two-stage training, \textit{DriveDreamer} has acquired a comprehensive understanding of the driving world, encompassing the structural constraints of traffic, predictions of future driving states, and interaction with the established world model.

\section{Experiment}
\subsection{Experiment Details}

\noindent
\textbf{Dataset.} The training data is sourced from the real-world driving dataset nuScenes \cite{nusc}, comprising a total of 700 training videos and 150 validation videos. Each video includes $\sim$20 seconds of footage captured by six surround-view cameras. The videos have a frame rate of 12Hz, resulting in $\sim$1M video frames available for training. During the first-stage training, we utilize the nuScenes-devkit \cite{nudev} to acquire HDMap annotations (lane boundary, lane divider, and pedestrian crossing) corresponding to 12Hz frames, which are then projected onto the image plane. Considering the nuScenes dataset only provides 2Hz 3D bounding box annotations, we supplement this with 12Hz bounding box annotations from \cite{asap}. In the second-stage training, we employ the yaw angle and velocity of the ego-car as the driving action inputs. Besides, we extract scene description information (\eg, weather and time) from the nuScenes annotation, which serves as text conditions.

\noindent
\textbf{Training.}
The proposed \textit{Auto-DM} is built upon Stable Diffusion v1.4 \cite{df6}, whose original parameters are frozen. In step 1 of first-stage training, our model is trained for 40 epochs with a batch size of 16. In step 2, \textit{Auto-DM} is trained for 10 epochs with a batch size of 1, with video frame length $N=32$, and spatial size of $448\times 256$. During second-stage video prediction training, our model predicts 16 frame driving videos $I_{1:16}$ and 16 future driving actions  $I_{17:32}$, and the model is trained for 10 epochs on a batch size of 1. All the experiments are conducted on A800 GPUs, and we use the AdamW optimizer \cite{adam} with a learning rate $5\times10^{-5}$.

\noindent 
\textbf{Evaluation.} We conducted a comprehensive evaluation of the proposed \textit{DriveDreamer}, employing both qualitative and quantitative assessments. We utilized frame-wise Fréchet Inception Distance (FID) \cite{fid} and Fréchet Video Distance (FVD) \cite{fvd} to evaluate the generation quality, where the evaluated image is resized to $448\times 256$. Besides, to verify the generated images enhance the training of driving perception methods, \textit{DriveDreamer} is evaluated through 3D object detection, with FCOS3D \cite{fcos3d} and BEVFusion \cite{bevfusion} as baseline methods. Furthermore, we test the performance of driving policy generation. Following the settings in \cite{stp3}, we evaluate output driving trajectories for future 3 seconds.

\subsection{Controllable Driving Video Generation}

The proposed \textit{DriveDreamer} exhibits a profound comprehension of driving scenarios, capable of controllably generating diverse driving videos. In this subsection, we first demonstrate that, based on first-stage training, \textit{DriveDreamer} can generate diverse driving videos under structured traffic conditions. Besides, we verify that the generated images can enhance the training of driving perception methods. Furthermore, \textit{DriveDreamer} showcases its versatility by responding to different input actions, allowing for the control of the vehicle's trajectory and consequently generating diverse driving videos.

As shown in Fig.~\ref{fig:fig0} and Fig~\ref{fig:fig1}, \textit{DriveDreamer} exhibits proficiency in producing images and videos that adhere meticulously to structured traffic conditions (more visualizations are in supplement). Significantly, we can also manipulate the text prompt to induce variations in the generated videos, encompassing changes in weather and time of day. 
To further validate the generation quality, we extract 4K traffic conditions (from the nuScenes training set) to generate driving images. The generated images are combined with real images for training the 3D detection task. Results in Tab.~\ref{tab:tab1} indicate that training with our synthetic data significantly enhances the performance of 3D detection. Specifically, compared with training without synthetic data, the mAP metrics of FCOS3D and BEVFusion are improved by 0.7 and 3.0.

In addition to the utilization of structured traffic conditions for generating driving videos, \textit{DriveDreamer} exhibits the capability to diversify the generated driving videos by adapting to different driving actions. As depicted in Fig.~\ref{fig:fig0} (more visualizations are in supplement), starting from an initial frame paired with its corresponding structural information, \textit{DriveDreamer} can generate distinct videos based on various driving actions, such as videos depicting left and right turns. In summary, \textit{DriveDreamer} excels in producing a wide spectrum of driving scene videos, characterized by both high controllability and diversity. Thus, \textit{DriveDreamer} holds promise for training autonomous-driving systems across a wide range of tasks, encompassing even corner cases and long-tail scenarios.

\begin{table}[t]
  \centering
  \resizebox{0.47\textwidth}{!}{
    \begin{tabular}{l c c|c c}
         \hline

    Methods & 
      Resolution  & 
      Data  &  mAP ($\uparrow$)  & NDS ($\uparrow$)\\

        \hline\hline
       \noalign{\smallskip}
          FCOS3D \cite{fcos3d} & $1600\times 900$ & w/o synthetic data&30.2& 38.1\\ 
           FCOS3D \cite{fcos3d} & $1600\times 900$ & w 4K synthetic data&\textbf{30.9} \textcolor[RGB]{0,133,21}{(+0.7)} & \textbf{38.3} \textcolor[RGB]{0,133,21}{(+0.2)}\\ 
           BEVFusion \cite{bevfusion} & $704\times 256$ & w/o synthetic data & 32.8& 37.6\\ 
           BEVFusion \cite{bevfusion} & $704\times 256$ & w 4K synthetic data &\textbf{35.8} \textcolor[RGB]{0,133,21}{(+3.0)} & \textbf{39.5} \textcolor[RGB]{0,133,21}{(+1.9)}\\ 
         \noalign{\smallskip}
         \hline
    \end{tabular}}
  \caption{Performance of synthetic data augmentation on training 3D object detection.}
    \label{tab:tab1} 
    \vspace{-1.5em}
\end{table}

In the quantitative experiment, we extract ego-car driving actions from the nuScenes validation set as conditions to generate driving videos. For comparison, we train DriveGAN \cite{drivegan} on the nuScenes dataset, employing the same training settings as those used for \textit{Drivedreamer}. Besides, we train \textit{Drivedreamer} without \textit{ActionFormer} as a baseline (specifically, the action features are directly concatenated with the zero-padded structured traffic conditions). The results are presented in Tab.~\ref{tab:tab2}, where we evaluate the quality of generated videos. Notably, our approach without first-stage training achieves superior FID and FVD scores compared to DriveGAN. This observation underscores the effectiveness of leveraging a powerful diffusion model in visually comprehending driving scenarios. Furthermore, our findings reveal that \textit{Drivedreamer} after first-stage training, exhibits an improved understanding of the structured information within driving scenes, resulting in higher-quality video generation. Lastly, we observe that the proposed \textit{ActionFormer} effectively leverages the traffic structural information knowledge acquired during the first-stage training. Compared to the \textit{concatenation} baseline approach, the \textit{ActionFormer}  iteratively updates future structured information based on input actions, which further enhances the quality of generated videos.

\begin{table}[t]
  \centering
  \resizebox{0.47\textwidth}{!}{
    \begin{tabular}{l|c|c|c|c}
         \hline

    \multirow{2}{*}{\text { Methods }} & 
      1st-stage train  & 
      2nd-stage train  &  
     \multirow{2}{*}{FID ($\downarrow$)} &   
     \multirow{2}{*}{FVD ($\downarrow$)} \\
     &(\textit{Auto-DM}) &(\textit{ActionFormer}) &\\

        \hline\hline
       \noalign{\smallskip}
          DriveGAN \cite{drivegan} &-&-& 27.8& 390.8\\ 
          \textit{DriveDreamer} &  &  & 15.9& 363.3\\ 
          \textit{DriveDreamer} & \checkmark & & 15.3& 349.6\\ 
          \textit{DriveDreamer} &\checkmark & \checkmark & \textbf{14.9} & \textbf{340.8}\\ 
         \noalign{\smallskip}
         \hline
    \end{tabular}}
  \caption{Comparison of generation quality on nuScenes validation.} 
    \label{tab:tab2} 
    \vspace{-1.5em}
\end{table}

\begin{table}[t]
  \centering
  \resizebox{0.48\textwidth}{!}{
    \begin{tabular}{l|c|c|c|c}
         \hline
          Method & Visual Info.& Action Info.&$L_2$  Avg. (m) &  Col. Avg. $(\%$)\\
        \hline\hline
       \noalign{\smallskip}
          ST-P3 \cite{stp3} & \checkmark& & 2.11&0.71 \\ 
          UniAD \cite{uniad} & \checkmark& & 1.65&0.31 \\ 
          AD-MLP \cite{zhai2023ADMLP} & &\checkmark & \textbf{0.29} & 0.19 \\ 
          VAD \cite{vad} & \checkmark& \checkmark& 0.37&\textbf{0.14} \\ 
          \textit{DriveDreamer} &\checkmark & \checkmark &\textbf{0.29} &0.15 \\ 
         \noalign{\smallskip}
         \hline
    \end{tabular}}
    \caption{Open-loop planning performance on nuScenes validation set. The evaluation settings  are the same as ST-P3 \cite{stp3}.} 
    \label{tab:tab3} 
\end{table}

\subsection{Driving Action Generation}

In addition to its capacity for generating highly controllable driving videos, \textit{DriveDreamer} demonstrates the ability to predict reasonable driving actions. As depicted in Fig.~\ref{fig:fig0}, provided with an initial frame condition and past driving actions, \textit{DriveDreamer} can generate future driving actions that align with real-world scenarios . Furthermore, we conduct a quantitative assessment of the prediction accuracy. Specifically, MLP layers \cite{zhai2023ADMLP} are utilized to encode past driving action information. Additionally,  multi-scale UNet features are pooled as visual cues. The two modality features are then concatenated to learn future driving trajectories (more implementation details are in supplement). The results of open-loop evaluation on the nuScenes dataset are presented in Tab.~\ref{tab:tab3}. Remarkably, the average $L_2$ trajectory error of \textit{DriveDreamer} is merely 0.29m, surpassing the performance of the multi-modality method VAD \cite{vad}. In addition, \textit{DriveDreamer} relatively decreases the average collision rate reported in \cite{zhai2023ADMLP} by 21\%, confirming that the visual features learned by \textit{DriveDreamer} contribute to end-to-end autonomous driving, thereby enhancing driving safety. 


\section{Discussion and Conclusion}
\textit{DriveDreamer} represents a significant advancement in the field of world modeling, particularly in the context of autonomous driving. By focusing on real-world driving scenarios and harnessing the power of the diffusion model, \textit{DriveDreamer} has demonstrated its ability to comprehend complex environments, generate high-quality driving videos, and formulate realistic driving policies. While prior research primarily concentrated on gaming or simulated environments, \textit{DriveDreamer} extends the boundaries of world modeling to encompass the intricacies of actual driving conditions. \textit{DriveDreamer} paves the way for future research in autonomous driving, emphasizing the importance of real-world representation for more accurate modeling and decision-making in this critical domain.  


{\small
\bibliographystyle{ieee_fullname}
\bibliography{PaperForReview}
}

\newpage

In the supplement materials, we first elaborate on the implementation details of \textit{DriveDreamer}, including model architecture and synthetic data training details. Then, we present additional visualization results.
\looseness=-1

\section{Implementation Details}

\noindent
\textbf{Condition encoders.}
In \textit{DriveDreamer}, diverse encoders are employed to embed different condition inputs, including the reference image, HDMap, 3D box, and action. The detailed architectures of these encoders are listed in Table~\ref{tab:tab4}. For spatially aligned conditions, such as the reference image $I\in\mathcal{R}^{H\times W\times 3}$ and HDMap $H\in\mathcal{R}^{H\times W\times 3}$, a stack of 2D convolution layers is utilized to perform downsampling, ensuring the final output dimensions align with those of the diffusion noise. For unstructured conditions like the 3D box $B\in\mathcal{R}^{N\times N_B\times 16}$ and action $A\in\mathcal{R}^{N\times 2}$, Multilayer Perceptron (MLP) layers are employed for encoding features.

\begin{table}[ht]
  \centering
  \resizebox{0.47\textwidth}{!}{
    \begin{tabular}{l c c}
         \hline
    Conditions & Layer Description  & Output Size \\
        \hline\hline
       \noalign{\smallskip}
          Ref. Img. (Step A)& Conv2D, $4\times 4$, S4 & $H/4 \times W/4 \times 4$ \\ 
          Ref. Img. (Step B)& Conv2D, $4\times 4$, S4 & $H/16 \times W/16 \times 4$ \\ 
          Ref. Img. (Step C) & Conv2D, $4\times 4$, S4 & $H/64 \times W/64 \times 8$ \\ 
        \hline\hline
          HDMap (Step A) & Conv2D, $4\times 4$, S4 & $H/4 \times W/4 \times 4$ \\ 
          HDMap (Step B) & Conv2D, $4\times 4$, S4 & $H/16 \times W/16 \times 4$ \\ 
          HDMap (Step C) & Conv2D, $4\times 4$, S4 & $H/64 \times W/64 \times 8$ \\ 
        \hline\hline
          3D Box (Step A) & FourierEmbedder \cite{nerf} & $N \times N_B \times 256 $ \\ 
          3D Box (Step B) & MLP &  $N \times N_B \times 512$ \\ 
          3D Box (Step C) & MLP & $N \times N_B \times 768$ \\ 
        \hline\hline
        Action (Step A) & MLP & $N \times 32$ \\ 
         Action (Step B) & MLP & $N \times 128 $ \\ 
         \noalign{\smallskip}
         \hline
    \end{tabular}}
  \caption{Encoder architecture details, where S denotes stride, and each convolution layer and MLP layer are followed by \textit{Sigmoid Linear Units} \cite{silu}.}
    \label{tab:tab4} 
\end{table}

\begin{figure}[ht]
\centering
\resizebox{0.85\linewidth}{!}{
\includegraphics{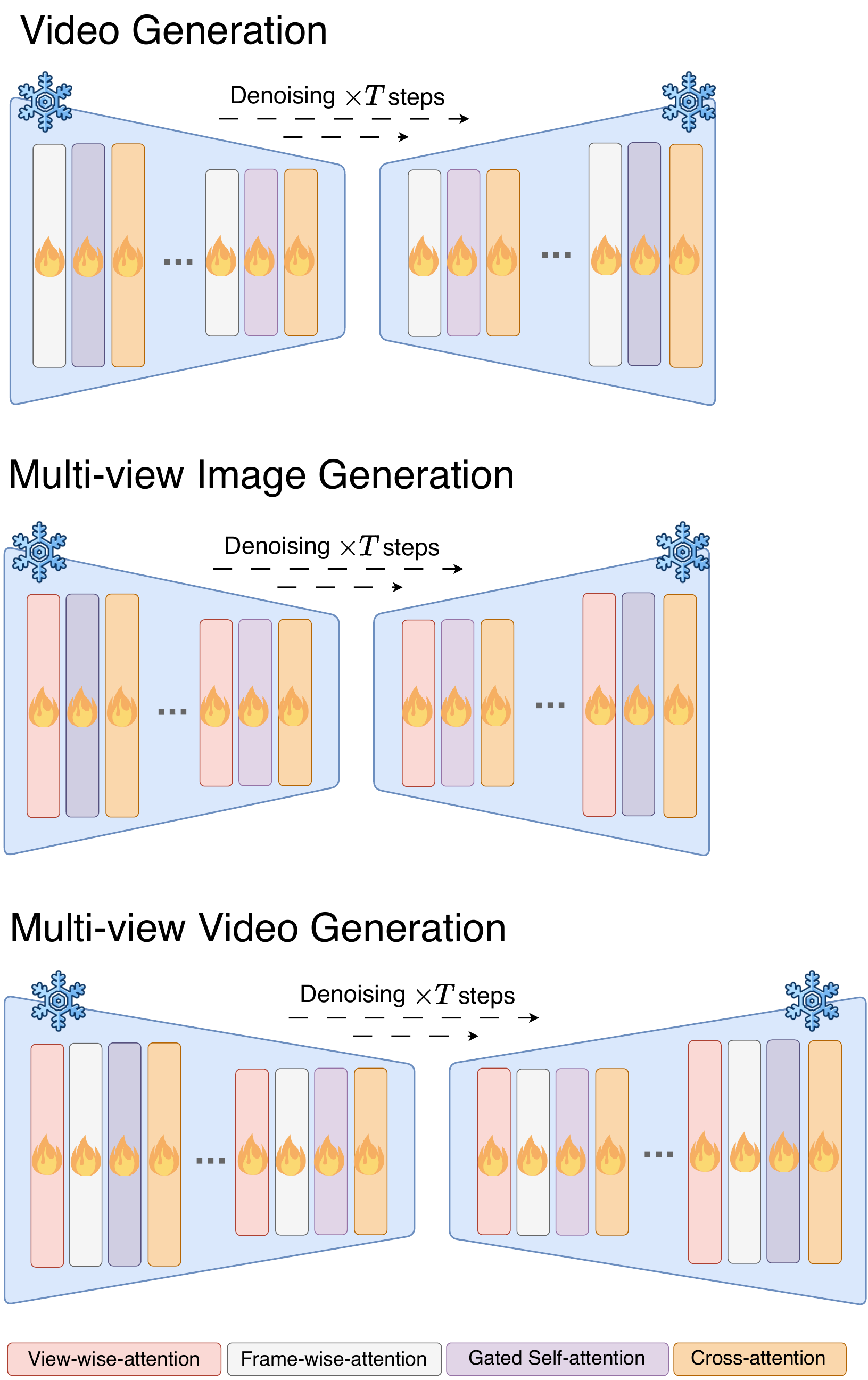}}
\caption{Model architecture comparison between video generation, multi-view image generation and multi-view video generation.}
\label{fig:mv-gen}
\end{figure}

\begin{figure*}[ht]
\centering
\resizebox{1\linewidth}{!}{
\includegraphics{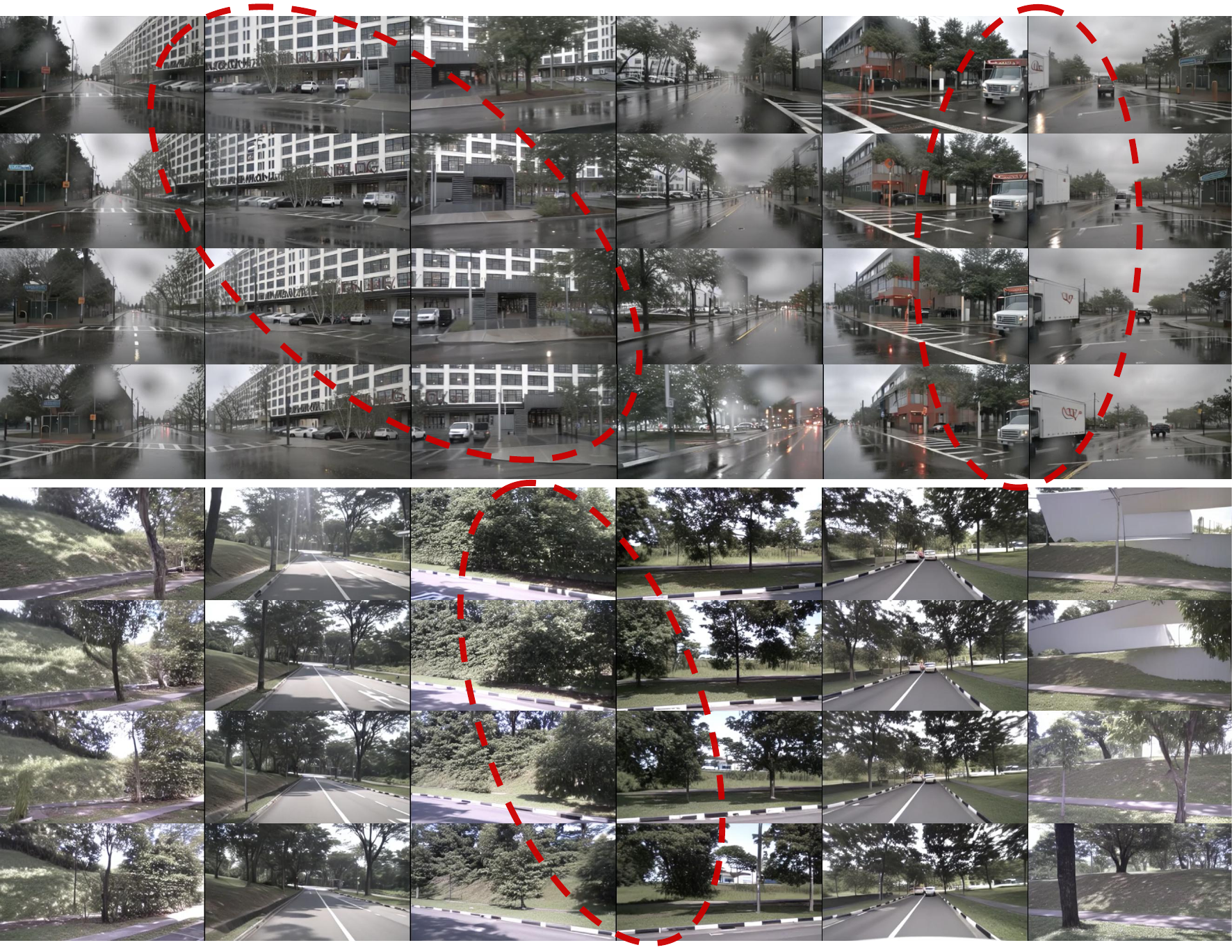}}
\caption{Visualizations of the generated multi-view video. Regions highlighted by \textcolor{red}{red} circles indicate that the generated videos are view-consistent and frame-consistent.}
\label{fig:mv}
\end{figure*}

\noindent
\textbf{Multi-view generation.} The framework of \textit{DriveDreamer} can be easily extended to multi-view image/video generation. The model architecture comparison between video generation, multi-view image generation and multi-view video generation are shown in Fig.~\ref{fig:mv-gen}. For multi-view image generation, the model framework is the same as that of video generation, except that the frame-vise attention layers are replaced with view-wise attention layers. Besides, the view-wise attention layers construct associations solely between adjacent views. For multi-view video generation, view-wise attention layers and frame-wise attention layers are stacked to process diffusion latent features, which results in view-consistent and frame-consistent videos (see Fig.~\ref{fig:mv}).

\noindent
\textbf{Action prediction architecture.} For action prediction, the multi-modal features are first concatenated:
\begin{equation}
    \text{CONCAT}(\mathcal{F}_{\text{p}}(U_0), \mathcal{F}_{\text{p}}(U_1), \mathcal{F}_{\text{p}}(U_2), \mathcal{F}_{\text{p}}(U_3), A_f),
\end{equation}
where $\mathcal{F}_{\text{p}}$ is the average pooling operation, $U_i(i=0,1,2,3)$ are multi-scale UNet features, and $A_f$ is the encoded driving action (i.e., velocity and yaw angle) features. Then we use MLP layers \cite{zhai2023ADMLP} to learn future driving actions. For trajectory prediction evaluation, following \cite{zhai2023ADMLP}, $A_f$ is additionally extracted from high-level command, accelerate and past trajectories, and we use the same action feature encoder of \cite{zhai2023ADMLP}.

\begin{figure*}[ht]
\centering
\resizebox{1\linewidth}{!}{
\includegraphics{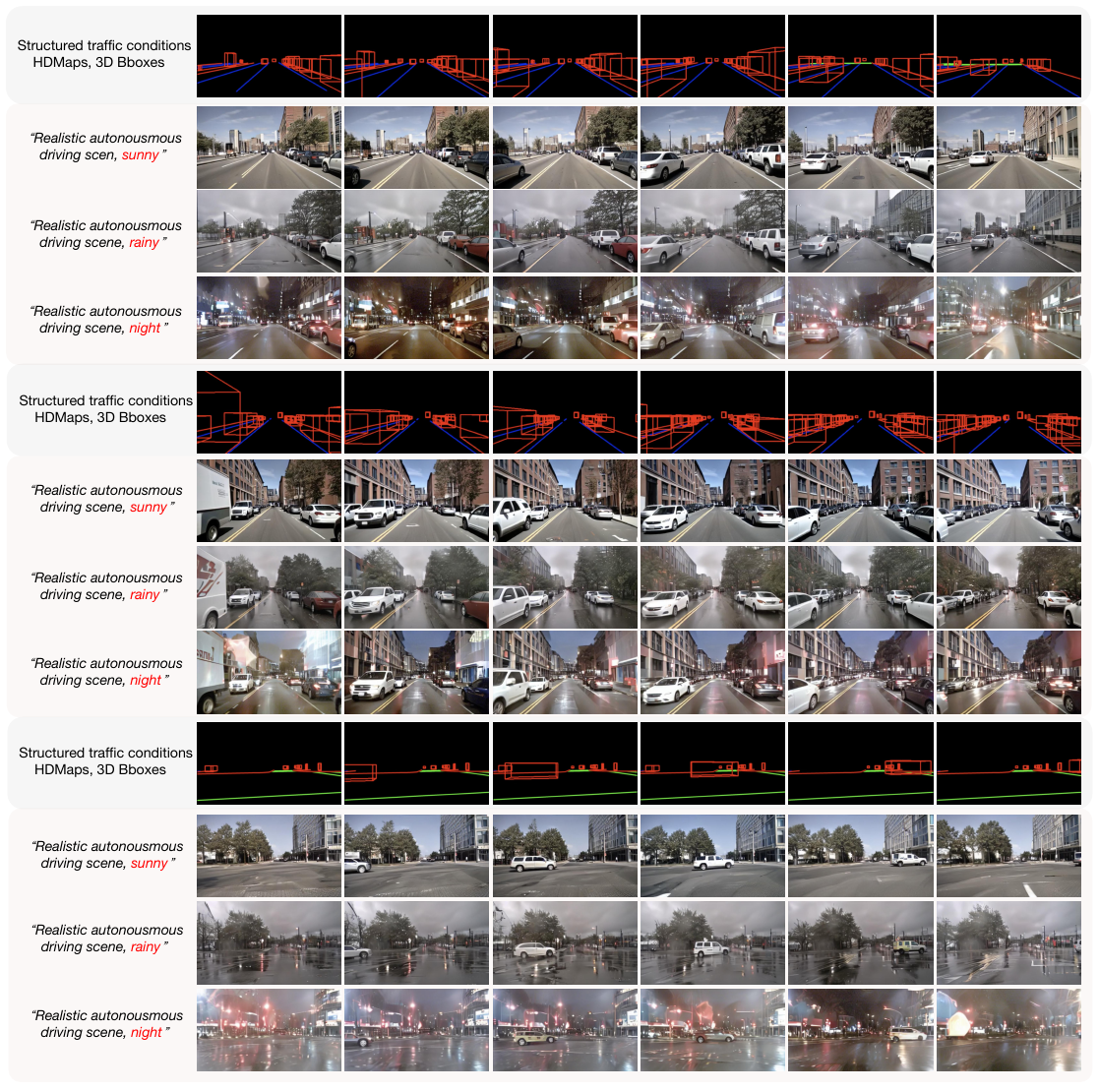}}
\caption{Driving video generation with structured traffic conditions (HDMaps and 3D boxes), where text prompts are utilized to adjust driving scenario style (\eg, weather and time of the day).}
\label{fig:s0}
\end{figure*}

\noindent
\textbf{Synthetic data training.} We leverage data generated by \textit{DriveDreamer} to augment the training of 3D detection tasks. Specifically, \textit{DriveDreamer} is fine-tuned with higher-resolution images, where the training data is from nuScenes \cite{nusc}. Consequently, \textit{DriveDreamer} can generate high-fidelity images with a resolution of $768\times 448$. Then the generated images are resized to the original resolution of $1600\times 900$, which can be utilized to train various off-the-shelf 3D detectors. During the training process, we randomly select 4000 samples (3D boxes and HDMap) from the nuScenes training set, which are employed to generate multi-view images. These synthetic data are mixed with the original training set to train 3D detectors. In the experiment, we train each baseline (i.e., FCOS3D \cite{fcos3d} and BEVFusion \cite{bevfusion}) for 12 epochs. The results presented in Tab.~\ref{tab:tab1} demonstrate that our approach significantly improves the performance of downstream tasks.

\section{Visualizations}
\begin{figure*}[t]
\centering
\resizebox{0.87\linewidth}{!}{
\includegraphics{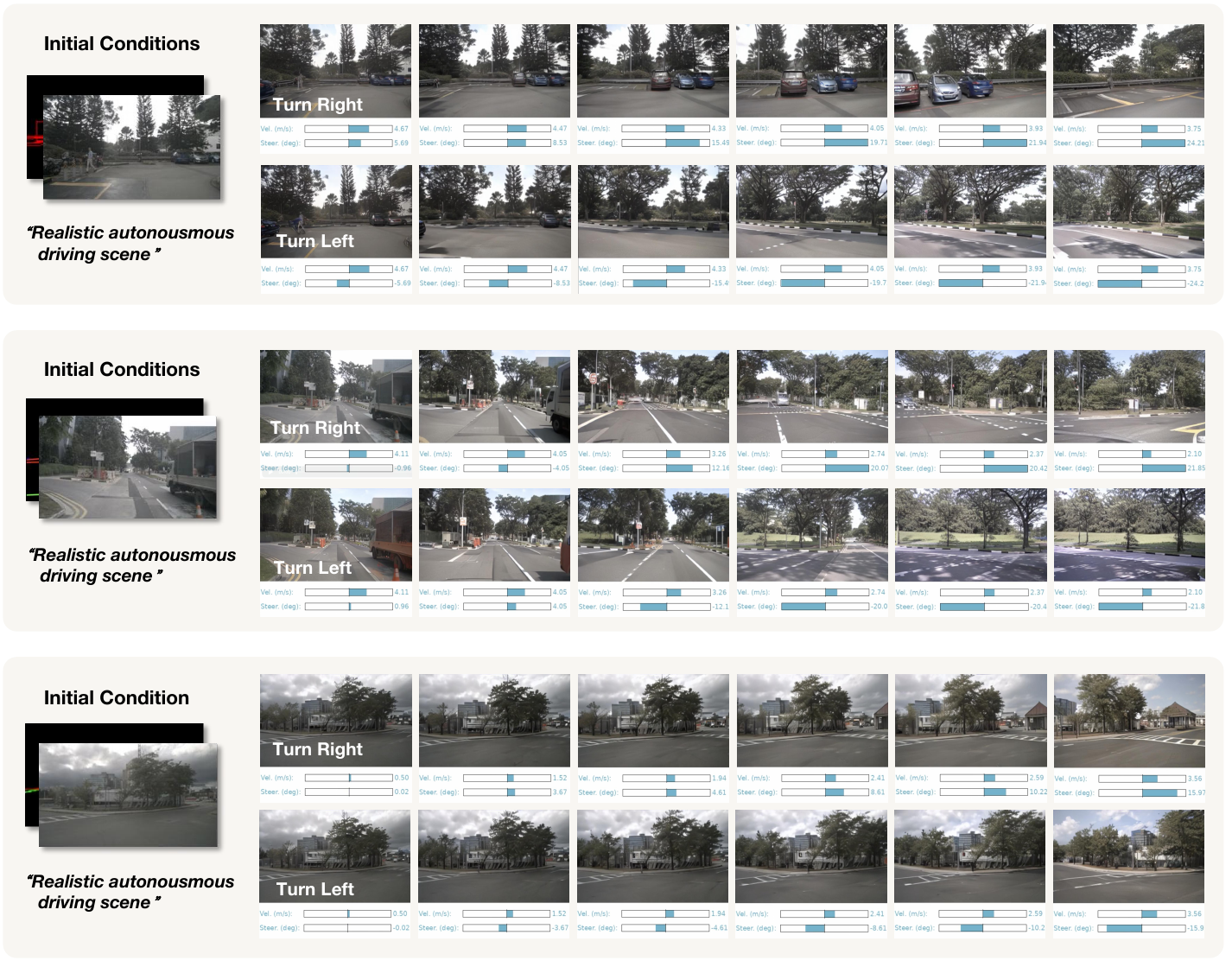}}
\caption{Future driving video generation with driving actions interaction, where different driving actions (\eg turn left, turn right) can produce corresponding driving videos.}
\label{fig:s1}
\end{figure*}

\begin{figure*}[t]
\centering
\resizebox{0.87\linewidth}{!}{
\includegraphics{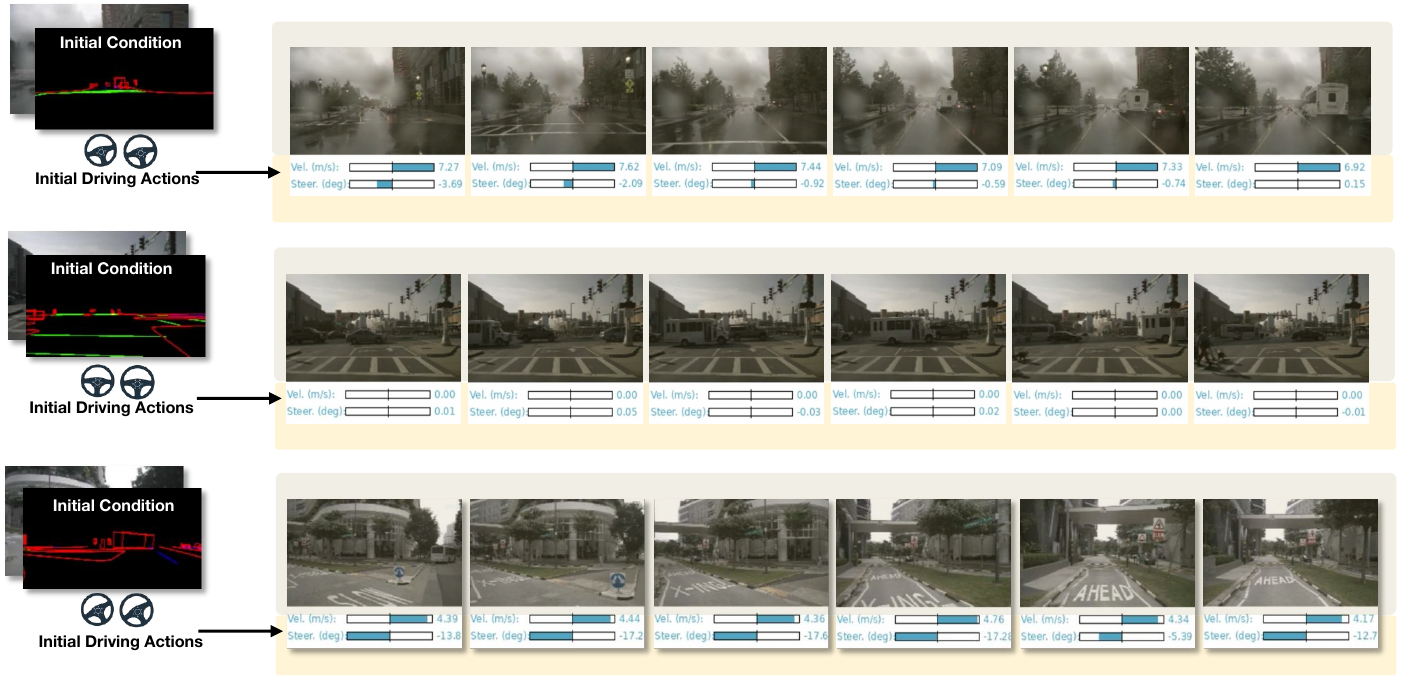}}
\caption{Visualization of the predicted future driving actions, along with the corresponding ground truth driving video.}
\label{fig:s2}
\end{figure*}

As shown in Fig.~\ref{fig:s0}, \textit{DriveDreamer} exhibits significant proficiency in producing a diverse range of driving scene videos that adhere meticulously to structured traffic conditions, comprising elements such as HDMaps and 3D boxes. Significantly, we can also manipulate the text prompt to induce variations in the generated videos, encompassing changes in weather and time of day. This heightened adaptability contributes substantially to the multifaceted nature of the generated video outputs.
In addition to the utilization of structured traffic conditions for generating driving videos, \textit{DriveDreamer} exhibits the capability to diversify the generated driving videos by adapting to different driving actions. As depicted in Fig.~\ref{fig:s1}, starting from an initial frame paired with its corresponding structural information, \textit{DriveDreamer} can generate distinct videos based on various driving actions, such as videos depicting left and right turns. Apart from its capacity for generating highly controllable driving videos, \textit{DriveDreamer} demonstrates the ability to predict reasonable driving actions. As depicted in Fig.~\ref{fig:s2}, provided with an initial frame condition and past driving actions, \textit{DriveDreamer} can generate future driving actions that align with real-world scenarios. Comparative analysis of the generated actions against corresponding ground truth videos reveals that \textit{DriveDreamer} consistently predicts sensible driving actions, even in complex situations such as intersections, obeying traffic lights, and executing turns.

\end{document}